# ARTIFICIAL NEURAL NETWORKS AND THEIR APPLICATIONS

Nitin Malik

Department of Electronics and Instrumentation Engineering,
Hindustan College of Science and Technology, Mathura, India.

*Abstract*: The Artificial Neural Network (ANN) is a functional imitation of simplified model of the biological neurons and their goal is to construct useful 'computers' for real-world problems and reproduce intelligent data evaluation techniques like pattern recognition, classification and generalization by using simple, distributed and robust processing units called artificial neurons. ANNs are fine-grained parallel implementation of non-linear static-dynamic systems. The intelligence of ANN and its capability to solve hard problems emerges from the high degree of connectivity that gives neurons its high computational power through its massive parallel-distributed structure. The current resurgent of interest in ANN is largely because ANN algorithms and architectures can be implemented in VLSI technology for real time applications. The number of ANN applications has increased dramatically in the last few years, fired by both theoretical and application successes in a variety of disciplines. This paper presents a survey of the research and explosive developments of many ANN-related applications. A brief overview of the ANN theory, models and applications is presented. Potential areas of applications are identified and future trend is discussed.

## I. INTRODUCTION

In recent years there has been a confluence of ideas and methodologies from several different disciplinary areas to give rise to an extremely interesting research area called Artificial Neural Network (ANN) [1]. A neuron is the fundamental building block of nervous system that performs computational and communication function. The ANN is a functional imitation of simplified model of the biological neurons and their goal is to reproduce intelligent data evaluation techniques like pattern recognition, classification and generalization by using simple, distributed and robust processing units called artificial neurons or Processing Elements (PE) [2]. The artificial neuron was designed to mimic the first-order characteristics of biological neurons. The intelligence of ANN and its capability to solve *hard problems* emerges from the high degree of connectivity that gives neurons its high computational power (processing capability) through its massive parallel-distributed structure or architecture, each neuron of which performs only very limited operation. Even though individual neuron works very slowly, they can still quickly find a solution by working in parallel.

A major advantage of ANN approach is that the domain knowledge is distributed in the neurons and information processing is carried out in parallel-distributed manner [3]. ANNs are highly parallel data processing tools capable of learning functional dependencies of data [2]. Being adaptive units they are able to learn these complex relationships even when no functional model exists. This provides the capability to do "Black Box Modeling" with little or no prior knowledge of the function itself. ANNs are fine-grained parallel implementation of non-linear static-dynamic systems. They have the ability to properly classify a highly non-linear relationship and once trained, they can classify new data much faster than it would be possible, by solving the model analytically.

The current resurgence of interest in ANNs is largely due to the emergence of powerful new methods as well as to the availability of computational power suitable for simulation. The field is particularly exciting today because ANN algorithms and architectures can be implemented in

VLSI technology for real-time applications. [4] Currently, two principal streams can be identified in ANN research. Researches in the first stream are concerned with modeling the brain and thereby explain its cognitive behaviour. On the other hand, the primary aim of researchers in the second streams is to construct useful 'computers' for real-world problems of pattern recognition by drawing these principles.

## II. ANN THEORY AND MODELS

Different type of Neural Networks (NN) has been proposed but all of them have three things in common: the individual neuron, the connection between them (architecture), and the learning algorithm. Each type restricts the kind of connections that are possible. For example, it may specify that if one neuron is connected to another, then the second neuron cannot have another connection towards the first. The type of connection possible is generally referred to as the *architecture* or the topology of the neural network.

A neural network consists of one or more layers of neurons. In large number of NN models, such as Perceptron, Linear Associator, Multi-layer feed-forward networks with Back-Propagation (BP) learning, the Boltzmann machine and the Grossberg model, the output from the units from one layer is only allowed to activate neurons in the next layer [5]. However in some models, such as Kohonen nets and Hopfield model the signal is allowed to activate neurons in the same layer. In models like Self- organising Feature Map (SOFM), the network connects a vector of inputs to a two-dimensional grid of output neurons [6, 7]. Figure 1. shows a general classification of ANN models.

A connection between a pair of neurons has an associated numerical strength called *synaptic weight* or adaptive coefficient [8]. The strength of interconnectivity can be represented as a weight matrix with positive (excitory), negative (inhibitory), or zero (no connection) values [5]. The weight determines the structure of the signal which is transmitted from one neuron to another thus coding the knowledge of the network. When the cumulative excitation exceeds the cumulative inhibition by an amount called threshold (T), typically a value of 40 mV, the neuron fires sending the signals down to other neurons. [9] Only some of the networks provide instantaneous response. Other networks need time to respond and are characterized by their time-domain behaviour, which we referred to as *neural dynamics.* The time interval between inputs are applied and neurons give output is called *period of latent summation.*

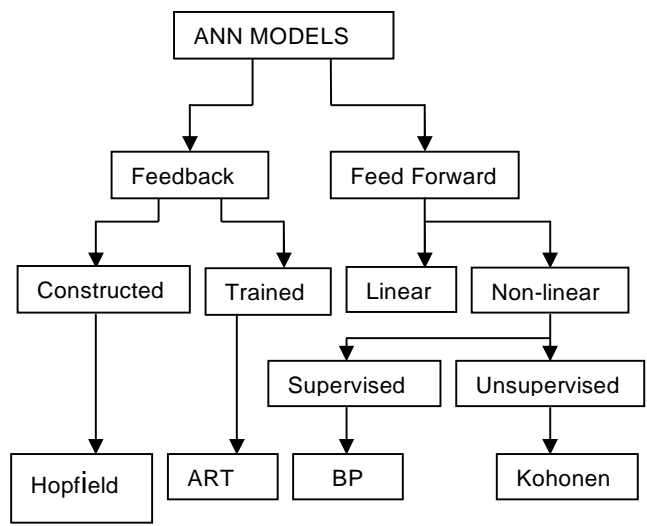

Fig. 1. General classification of ANN models

A neuron is said to be 'trainable' if its threshold and input weights are modifiable. Inputs are presented to the neurons. If the neurons does not give the desired output (determined by us), then it had made a mistake. Then some weights and thresholds have to be changed to compensate for the error. The rules which govern how exactly these changes are to take place is called *learning (or training) algorithm*. Learning algorithms differ from each other in a way in which the adjustment to synaptic weights of a neuron is formulated.

The weights of the network are incrementally adjusted so as to improve a predefined *performance measure* over time. The learning process is best viewed as "search" in a multi-dimensional weight space for a solution, which gradually optimizes a pre-specified objective function. The NN becomes more knowledgeable about its environment after each iteration of the learning process.

In order for the net to learn one need to present a number of examples to the net whose attributes are known or are representatives for the unknown model [2]. The set of given examples is called the training set or training patterns.

After the *training* period, the network should be able to give correct output for any kind of input. This is called *testing*. If it was not trained for that input, then it should try to give reasonable output depending on how it was trained. This is called *generalization*. The actual method of determining the output for a given set of inputs is called the processing algorithm [2]. Different NNs are characterized by difference in the architecture, the learning algorithm, and the processing algorithm.

Every learning algorithm contains basically a learning rule. There are two main rules available for learning: *Hebbian rule* for *supervised learning* and D*elta rule* for *unsupervised learning*. Adaptation of these by simple modifications to suit a particular context generates in many other learning rules. Supervised learning requires the pairing of each input vector with the target vector representing the desired output, together these are called training pairs [10]. The desired output represents the optimum action to be performed by the NN. [8] Supervised NN may be feed-forward network such as Multi-Layer Perceptron (MLP), Functional Link Network (FLN), Radial Basis Function Network (RBFN), Parallel Self-Organising Hierarchal Neural Network (PSHNN), or a feed-back network such as Hopfield network. Unsupervised learning requires no target vectors for the outputs. The learning algorithm modifies network weights in response to the inputs to produce output vectors that are consistent. Kohonen's SOFM and Adaptive Resonance Theory (ART) are the examples of unsupervised NN.

Unlike *Expert Systems*, NN do not give an explicit set of rules that match the input it receives to the output it is told correct (input-output mapping). This ability to learn by examples is the characteristic of the ANN. Thus they can modify their behaviour in response to their environment. Figure 2. shows an artificial neuron. The input to the neuron can be from the actual environment or from the other neurons. Its output can be fed into other neurons or directly into the environment [2]. The OUT (output of the neuron) is constructed by taking the *weighted sum* of the inputs [called NET (net input to a neuron) or *combination function* (vector-to-scalar function)] transformed by transfer function F [also called activation function (scalar-to-scalar function)]. This transfer function introduces non-linearity into the system. This makes the system so powerful.

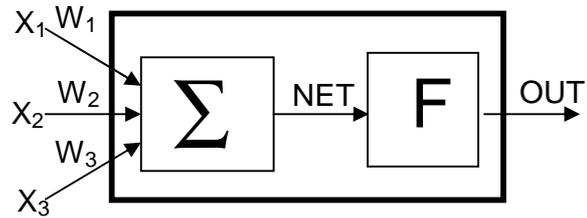

Fig. 2. Artificial Neuron

The net input to the *i*th *neuron is given by*

$$NET_i = X_1W_1 + X_2W_2 + \ldots X_nW_n$$

$$NET_i = \sum_i W_i X_i$$

The output of the *i*th neuron is given by
$$OUT_i = F(NET_i)$$

$$OUT = \begin{cases} 1 & if \quad NET \geq T \\ 0 & if \quad NET < T \end{cases}$$

### III. ANN APPLICATIONS

Since their introduction, ANNs have found many applications in many of the areas. Table 1. shows the generic applications of various ANN models in chronological sequence. The applications given below are few among many reported in the literature:

*Alarm Processing*
In emergencies, engineers are expected to quickly evaluate various options and implement an optimal corrective action. However, the number of real-time messages (alarms) received on the VDUs is too large for the time available for their evaluation. Processing such alarms in real-time and alerting the operator to the root cause or the most important of these alarms has been identified as a valuable operational aid. ANNs have been implemented for such alarm processing. Chan [11] proposed a three-layer Perceptron network for this purpose, with promising simulation results. The fast response of a trained ANN and its generalization abilities become very useful in this application.

Table1. Generic applications of various ANN models

| Application Area | ANN model used | Year |
| --- | --- | --- |
| Typed character recognition | Perceptron | 1957 |
| Echo cancellers in telephone lines | Madaline | 1960 |
| Speech recognition | Avalanche | 1967 |
| Speech synthesis | Back Propagation | 1974 |
| Knowledge extraction from databases | Brain state in a box | 1977 |
| Handwritten character recognition | Neocognition | 1978 |
| RADAR and SONAR identification | ART | 1978 |
| Aircraft navigation | Self-Organising map | 1980 |
| Image processing | Hopfield | 1982 |
| Associative memories | Hopfield | 1985 |
| Loan application evaluation | Counter Propagation | 1986 |

*Eddy current analysis*

Analysis of eddy current losses requires numerical solution of Integra-differential equations. Discretising these equations and solving them using finite-element methods is computationally very expensive. Feria *et al.* [12] *report* a cellular ANN which produced a faster, computationally less expensive and simpler method of solving these equations. They proposed a cellular NN as an alternative to finite-element methods. The cellular networks were simulated using SPICE. The cellular network calculated eddy currents and eddy current losses in a source current carrying conductor in a time-varying magnetic field. This implementation opens up a wide range of applications in structural analysis, electromagnetic field computations, etc.

*Harmonic source monitoring*

Hartana and Richards [29, 30] report identification and monitoring of harmonic sources in the systems containing non-linear loads. This approach assumes sufficient direct measurement of harmonics in the system. These authors employed multiple three-layer Perceptrons. The ANNs were trained using simulation results for varying load conditions. The ANNs were used in conjunction with a state estimator to pinpoint and monitor the source of the harmonics. This approach was able to identify a harmonic source which had not been identified previously.

*Applications in nuclear power plants*

In a project initiated by US Department of Energy, researchers at the University of Tennessee have investigated the potential applications of ANNs in enhancing the safety and efficiency of nuclear power plants [15, 16, 17]. The areas under investigation were: diagnosis of specific abnormal conditions, detection of the change of mode of operation, signal validation, monitoring of check valves, modeling of the plant thermodynamics, monitoring of plant parameters, analysis of plant vibrations, etc.

## IV. CONLUSIONS

The ANN has an ability to develop a generalized solution to the problem other than that used for training and to produce valid solutions, even when there are errors in the training data. These factors combine to make NN a power tool for modeling the problems in which functional relationships are subject to uncertainty or likely to vary with the passage of time.

Another area that can be benefited from NN approach is where the time required to generate solution is critical such as real-time applications that require many solutions in quick succession. The ability of a NN to produce quick solutions irrespective of the complexity of the problem makes them valuable even when alternative techniques are available that can produce more accurate solutions.

[2] In contrast to classical techniques or expert systems, which attempts to formalize knowledge and develop partial qualitative system models, ANN do not provide a formal representation of the relation between input and output data. It is therefore necessary to check the ANN performance by stastical tests.

## V. FUTURE OUTLOOK

Each type of NN has been designed to tackle a certain class of problems. Hopefully, at some stage we will be able to combine all the types of NNs into a uniform framework. Hopefully, we will reach our goal of combining brains and computers.

Basically ANN is a part of *Artificial Intelligence* (AI). The success so far has been in the simulation of intelligence- not the creation of true intelligence. Therefore, ANN may become the foundation for more intelligent systems. Since ANN can be viewed as "low level" data processing tool, *hybrid approaches*, that are a combination of ANN with other techniques like Expert Systems, Fuzzy Logic, and Genetic Algorithms (GA), are promising areas to be investigated. Presently, integration of Fuzzy logic with ANN is a major area of research as it combines the advantage of both these fields [18]. GA have been increasingly applied in ANN design such as topology and parameter optimization.

**Nitin Malik** received his M.E. degree in electrical engineering from Madhav Institute of Technology and Science (M.I.T.S.), Gwalior, India. Currently he is a lecturer at Hindustan College of Science and Technology, Mathura, India. His areas of interests are power system security analysis, optimization and control, and fuzzy neural applications to power system.